\documentclass{article}

\usepackage[preprint]{neurips_2026}
\makeatletter
\renewcommand{\@noticestring}{}
\makeatother
\usepackage[utf8]{inputenc}
\usepackage[T1]{fontenc}
\usepackage{hyperref}
\usepackage{url}
\usepackage{booktabs}
\usepackage{graphicx}
\usepackage{amsfonts}
\usepackage{amsmath}
\usepackage{nicefrac}
\usepackage{microtype}
\usepackage{xcolor}

\workshoptitle{E-Values: From Statistics to ML}
\title{Discovering Translation-Worthy Languages with E-Values}
\author{
Wajdi Ben Saad \\
Carthago Labs \\
Paris, France \\
\texttt{wajdi@carthagolabs.org}
\And
Safa Madiouni \\
IÉSEG \\
Paris, France \\
\texttt{s.madiouni@ieseg.fr}
}

\begin{document}
\maketitle

\begin{abstract}
Choosing when to translate multilingual documents is a central routing problem in
text classification: translation can improve predictions for some languages
while degrading others or adding unnecessary computation. Uniform translation
and heuristic language tiers do not provide statistically controlled route
selection. We introduce a language-level router based on paired e-processes
that continuously compares direct and translation-assisted classification
before freezing a routing policy. A familywise-controlled threshold of 280
bounds the probability of any false route across 14 eligible languages per
dataset by 0.05. On SIB-200 and MASSIVE, the router selects translation for 4 of
15 languages and 14 of 15 locales, improving held-out accuracy over direct
classification by 8.14 and 16.70 percentage points, respectively. All 28
decisions remain stable across 50 outcome-independent orderings and relative to
the per-group threshold. Our results demonstrate that paired e-processes enable
statistically controlled, anytime-valid, and auditable multilingual
classification routing.
\end{abstract}

\section{Introduction and related work}
\label{sec:introduction}

Inference systems increasingly choose between a default computation and a
costlier intervention. When evidence is monitored as it accumulates,
fixed-horizon tests do not justify data-dependent stopping. E-values are
nonnegative evidence measures with null expectation at most one, and e-processes
extend this principle to continuous monitoring under explicit sequential
assumptions \citep{vovk2021evalues,grunwald2024safe,ramdas2023gametheoretic}.
We study a new
decision role for this established methodology: constructing a persistent
group-level router from paired outcomes.

Multilingual classification provides a concrete application. Inputs may be
classified directly by a multilingual model or translated into English before
classification. Translate-test performance varies with language, task, and
pipeline \citep{artetxe2023revisiting}, yet existing language-level routing uses
hand-specified resource tiers \citep{bensaad2026routing}. Such tiers neither
measure evidence that translation improves a group nor necessarily transfer
across tasks. Recent multilingual and translation-routing systems learn
instance-level strategy or model selection
\citep{wu2026noonefitsall,luo2026routelmt}, while FrugalML and RouteLLM learn
query-specific quality--cost policies \citep{chen2020frugalml,ong2025routellm};
our aim is different: assign a predefined group to the intervention only after
paired held-out observations provide sufficient statistical evidence.

Our construction follows the discordance principle behind McNemar's paired test
\citep{mcnemar1947}. Among examples on which the two paths disagree in
correctness, translation has higher marginal accuracy exactly when fixes are
more likely than regressions. We monitor this Bernoulli sequence with an
established mixture e-process for paired model evaluation
\citep{choe2023comparing,kotawala2026resolution}. PACE uses paired anytime-valid
evidence for sequential agent-update acceptance; here, crossings define a
persistent group-level inference route \citep{shawn2026pace}. All routes are
frozen before test evaluation.

We evaluate independently discovered routers on SIB-200 topic classification
\citep{adelani2024sib200} and MASSIVE intent classification
\citep{fitzgerald2023massive}. Our contribution is not a new e-process, but
(i) a paired-evidence formulation of group routing, (ii) a discovery--freeze--test
protocol that turns crossings into an auditable policy, and (iii) evidence that
the resulting boundary is task-specific rather than a universal language tier.

\section{Method}
\label{sec:method}

\subsection{Sign-sufficient paired reduction}
For group $g$, let $D$ denote a default path and $T$ an intervention. On held-out
discovery example $i$, define correctness indicators
\begin{equation}
C^D_{gi}=\mathbf 1\{\widehat y^D_{gi}=y_{gi}\},\qquad
C^T_{gi}=\mathbf 1\{\widehat y^T_{gi}=y_{gi}\}.
\end{equation}
A discordant pair is a \emph{fix}, $(C^D,C^T)=(0,1)$, or a \emph{regression},
$(1,0)$. Write their probabilities as $a_g$ and $b_g$, let
$d_g=a_g+b_g$ be the discordance rate, and let $q_g=a_g/d_g$. Then
\begin{equation}
\Delta_{\mathrm{acc},g}=\mathrm{Acc}_{T,g}-\mathrm{Acc}_{D,g}
=a_g-b_g=d_g(2q_g-1).
\label{eq:delta}
\end{equation}
For $d_g>0$, $q_g>1/2$ if and only if $T$ has higher marginal accuracy.
Concordant pairs contribute equally to both accuracies, so discarding them loses
no information about the sign of the paired accuracy difference. If $d_g=0$,
the difference is zero and the group retains $D$. The target is paired accuracy,
not Macro-F1, whose nonlinear dependence on the multiclass confusion matrix
requires a different process.

\subsection{Discordance e-process}
In a prespecified order, let $X_{g,t}=1$ for the $t$th fix and $0$ for the $t$th
regression, and $S_{g,t}=\sum_{j=1}^tX_{g,j}$. We test
$H_{0,g}:q_g\leq1/2$ with the normalized uniform mixture
\begin{equation}
E_{g,t}=2\int_{1/2}^{1}
\frac{p^{S_{g,t}}(1-p)^{t-S_{g,t}}}{(1/2)^t}\,dp,
\qquad E_{g,0}=1.
\label{eq:eprocess}
\end{equation}
The factor two normalizes a uniform mixing distribution on $[1/2,1]$; this
mixture is fixed before observing outcomes and requires no effect-size tuning.
Let $\mathcal F_{g,t-1}$ contain the prespecified order and the first $t-1$
discordant outcomes. Under i.i.d. paired sampling, the ordered discordant
outcomes are Bernoulli with parameter $q_g$. More generally, we assume
$P(X_{g,t}=1\mid\mathcal F_{g,t-1})\leq1/2$ under the null. For each fixed
$p>1/2$, this condition makes the corresponding likelihood-ratio process a
nonnegative supermartingale; mixing over $p$ preserves the e-process property
\citep{vovk2021evalues,ramdas2023gametheoretic}.

\subsection{Anytime validity and evidence allocation}
For a prespecified testing level $\alpha_g$, define the first crossing
$\tau_g=\inf\{t:E_{g,t}\geq1/\alpha_g\}$. The e-process maximal inequality gives
\begin{equation}
\Pr_{H_{0,g}}\!\left(\tau_g<\infty\right)
=\Pr_{H_{0,g}}\!\left(\sup_t E_{g,t}\geq1/\alpha_g\right)\leq\alpha_g.
\label{eq:anytime}
\end{equation}
Thus the same threshold is valid under continuous inspection and stopping at
the first crossing. Because the router makes simultaneous decisions for 14
translation-eligible groups within each dataset, we use the familywise error
allocation $\alpha_g=.05/14$. The resulting e-value threshold is
$1/\alpha_g=280$, which guarantees by the union bound that the probability of
at least one false route is at most .05, without requiring independence across
groups. We retain the per-group allocation $\alpha_g=.05$ (threshold 20) as a
sensitivity analysis.

\subsection{Evidence time, routing, and freezing}
The process advances in discordance time, whereas data arrive in example time.
If a route crosses after $t$ discordances, approximately $t/d_g$ paired examples
are required; hence evidence requirements reflect both intervention advantage
$q_g$ and the frequency $d_g$ of informative comparisons. This quantity
describes potential monitoring efficiency rather than realized compute savings.
A hash of a fixed seed and example identifier fixes order before outcomes are
examined. Each group follows the same auditable sequence: generate paired
predictions, update only at discordances, select $T$ at the first crossing, and
otherwise retain $D$. The resulting route, threshold, group set, models, and
backends are frozen before test evaluation; test outcomes cannot revise them.

\section{Empirical evaluation}
\label{sec:evaluation}

\subsection{Setting}
We use 15 languages on seven-class SIB-200, with 800 discovery and 204 test
examples per language, and 15 locales on 60-class MASSIVE, with 2,033 discovery
and 2,974 test examples per locale. Each dataset has 14 translation-eligible
non-English groups; English is fixed to the direct path. The direct path uses multilingual MiniLM; the intervention
uses pinned OPUS-MT or NLLB translation and English MiniLM. Neither classifier
is task-fine-tuned. We compare direct-only, translation-only, the fixed tier of
\citet{bensaad2026routing}, and an exact one-sided paired-test router. Accuracy
is the inferential target; Macro-F1, translation rate, and latency describe the
resulting policy. Full configurations, per-language effects, confidence
intervals, and backend details are deferred to the supplement.
Conditional on the frozen routing decisions, policy-level accuracy intervals
use 2,000 paired bootstrap replicates with test examples resampled within
language. Code, configurations, and reproduction scripts are available in the
public repository.\footnote{\url{https://github.com/WajdiBenSaad/NeurIPS2026_E-Values}}

\subsection{Operating-characteristic analysis}
Separately from model evaluation, we simulate Bernoulli discordance streams at
null and alternative values of $q_g$. We use 10,000 repetitions, monitoring
budgets of 50, 100, 250, and 500 discordances, using the primary threshold 280
and per-group sensitivity threshold 20. We
also simulate 5,000 routers under independent null streams using each dataset's
observed group-specific discordance horizons. These analyses use no test
outcomes or additional model inference.

\begin{figure}[t]
\centering
\includegraphics[width=.495\textwidth]{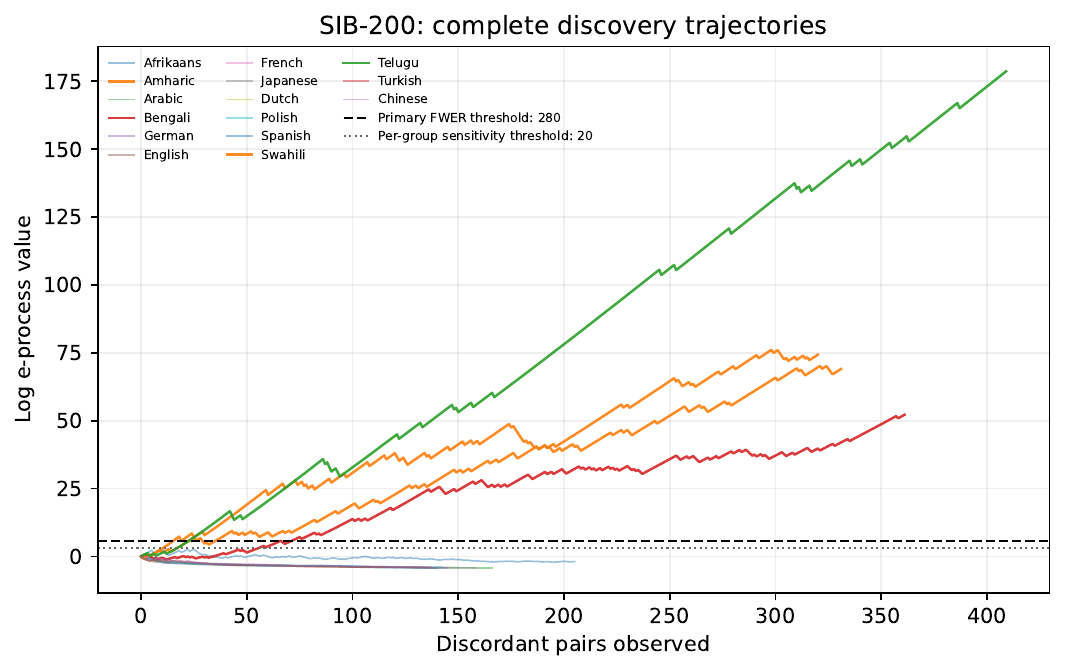}\hfill
\includegraphics[width=.495\textwidth]{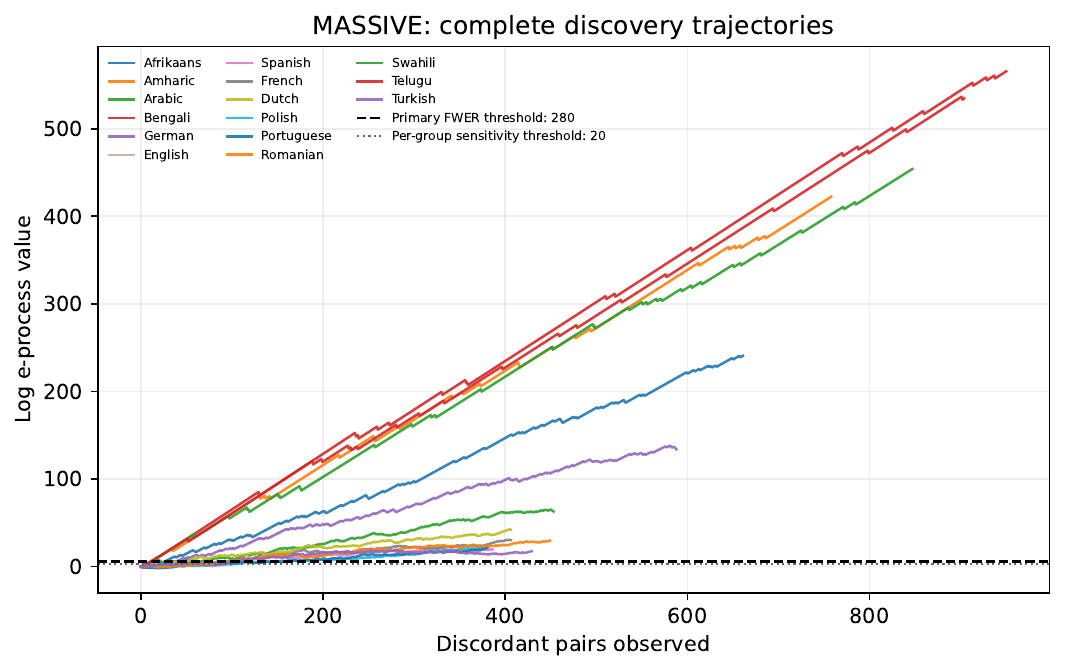}
\caption{Discovery e-processes for SIB-200 (left) and MASSIVE (right). The
primary familywise-controlled threshold is 280; threshold 20 is shown only as
the per-group $\alpha_g=.05$ sensitivity threshold. Routing uses the first
crossing, not the terminal or maximum value.}
\label{fig:trajectories}
\end{figure}

\subsection{Discovery evidence}
SIB-200 selects Swahili, Bengali, Telugu, and Amharic. At the primary threshold
280, their first crossings occur after 23, 66, 23, and 16 discordances,
respectively, and the same four
routes are already selected at a budget of 200 examples per language. Afrikaans
is the closest rejection: translation gains 1.13 discovery accuracy points and
its running maximum reaches 18.51, below both thresholds. MASSIVE
selects all 14 eligible locales, with primary-threshold crossings from 11
discordances for several locales to 141 for Polish
(Figure~\ref{fig:trajectories}). This contrast demonstrates task-dependent
routing in the two evaluated settings: the evidence boundary belongs to the
task and inference paths.

\subsection{Calibration, power, and evidence requirements}
At the boundary null $q=.5$, the empirical probability of crossing the primary
threshold 280 by 500 discordances is 0.26\%. Power rises from 9.50\% at $q=.55$
to 78.76\% at $q=.60$ and 99.87\% at $q=.65$. At the per-group sensitivity
threshold 20, corresponding power is 33.33\%, 94.84\%, and effectively one,
with a boundary-null crossing rate of 3.34\% (95\% CI: 3.01--3.71\%). Thus
familywise control requires more evidence, while power remains high at $q=.60$.
Table~\ref{tab:operating} reports the null operating points; the
full power curves are in Appendix~\ref{sec:appendix}.

\begin{table}[t]
\caption{Monte Carlo null crossing rates (\%). Threshold 280 is the primary
familywise-controlled decision threshold; threshold 20 is the per-group
$\alpha_g=.05$ sensitivity threshold. The per-stream column uses a
500-discordance horizon; router columns use 14 independent null streams and
observed group-specific horizons.}
\label{tab:operating}
\centering
\small
\setlength{\tabcolsep}{4.5pt}
\begin{tabular}{lrrrr}
\toprule
Rule & Threshold & Per stream & SIB router & MASSIVE router \\
\midrule
Primary familywise $\alpha_g=.05/14$ & 280 & 0.26 & 2.84 & 3.36 \\
Per-group sensitivity $\alpha_g=.05$ & 20 & 3.34 & 37.32 & 38.98 \\
\bottomrule
\end{tabular}
\end{table}

\subsection{Held-out confirmation}
All 28 frozen decisions agree with the direction of the held-out confirmatory test accuracy
effect. On SIB-200, translation raises accuracy by 23.04--42.16 points for the
four selected languages and has a negative held-out point estimate for every
rejected non-English language. The
e-router therefore translates 26.67\% of examples and obtains 74.05\% accuracy,
an 8.14-point gain over direct-only (95\% CI: 6.96--9.25), and slightly above the fixed tier while
translating one fewer language. On MASSIVE,
translation improves test accuracy for every eligible locale by 4.94--42.77
points. The e-router coincides with translation-only, consistent with positive
held-out accuracy point estimates for all eligible locales.
Relative to direct-only, its accuracy gain is 16.70 points (95\% CI:
16.28--17.14).
This provides no translation saving, but confirms that broad intervention is
supported for that task.
On SIB-200, selective routing reduces mean latency from 1.176 s for
translation-only to 0.347 s per example (70.5\% lower); on MASSIVE, it matches
translation-only at 0.372 s because all eligible locales are translated.

\subsection{Robustness and interpretation}
Importantly, all observed routing decisions are unchanged when moving from the
per-group threshold 20 to the familywise-controlled threshold 280. Thus, the
stronger router-wide error guarantee does not alter the empirical policy
selected on either dataset. The primary rule yields modeled router-wide null
crossing rates of 2.84--3.36\%, compared with 37.32--38.98\% in the per-group
sensitivity analysis.
Routes are also unchanged for $\alpha\in\{.01,.025,.05\}$ and across 50
outcome-independent orderings, including the primary ordering. The exact
paired-test router selects
the same groups; the contribution is therefore not superior endpoint selection,
but valid continuous monitoring, first-crossing evidence, and conversion of
that evidence into a reproducible frozen policy.

\section{Conclusion and limitations}
\label{sec:limitations}

The present process targets a positive paired-accuracy effect for each
group--task population; it does not yet encode a practical improvement margin
or intervention cost. The measured effects belong to the complete
translation-and-classification path.
Selective translation improves computational efficiency but may reinforce
performance disparities across linguistic groups.

Within these limits, paired e-processes replace fixed group heuristics with
auditable, task-specific routing decisions while keeping statistical evidence
distinct from downstream utility.

\appendix
\section{Technical appendices and supplementary material}
\label{sec:appendix}

The supplement contains frozen policy results, detailed latency summaries,
policy-level paired-bootstrap accuracy intervals, a synthetic threshold-crossing
power curve, a confirmatory quality--translation trade-off, and discovery-budget
sensitivity; the NeurIPS paper checklist follows. These materials use validated
stored predictions or synthetic discordance streams; no additional model
inference or translation was performed.

\paragraph{Compute resources.}
All experiments ran locally on a single MacBook Pro (MacBookPro16,1) with a
2.6~GHz six-core Intel Core i7, 16~GB RAM, and an AMD Radeon Pro 5300M GPU with
4~GB VRAM. The automatic device setting selected PyTorch MPS acceleration.
Encoder and translation batch sizes were 128 and 8, respectively, with one data
worker; languages were processed sequentially. Summed per-language elapsed
times were 3.39~h and 1.09~h for SIB-200 discovery and evaluation, and 2.83~h
and 4.72~h for MASSIVE discovery and evaluation, totaling 12.04~h for the final
runs. These figures exclude manual pauses between language launches and
exploratory or smoke-test runs. Subsequent statistical analyses reused cached
predictions and required no model inference or translation.

\begin{table}[ht]
\caption{Frozen application-level policy results. Translation and accuracy are
percentages.}
\label{tab:policies}
\centering
\small
\setlength{\tabcolsep}{5pt}
\begin{tabular}{llrrr}
\toprule
Dataset & Policy & Accuracy & Macro-F1 & Translation \\
\midrule
SIB-200 & Direct & 65.92 & .6655 & 0.00 \\
& Fixed tier & 73.82 & .7387 & 33.33 \\
& E-router & \textbf{74.05} & \textbf{.7407} & 26.67 \\
& Translate & 69.25 & .6963 & 93.33 \\
\midrule
MASSIVE & Direct & 27.36 & .3041 & 0.00 \\
& Fixed tier & 39.30 & .3627 & 33.33 \\
& E-router & \textbf{44.07} & \textbf{.3694} & 93.33 \\
& Translate & \textbf{44.07} & \textbf{.3694} & 93.33 \\
\bottomrule
\end{tabular}
\end{table}

\begin{table}[ht]
\caption{Measured policy latency in seconds per example. Times are assembled
from steady-state complete-path measurements and exclude model loading and cold
starts; they are hardware-specific.}
\label{tab:policy-latency}
\centering
\small
\setlength{\tabcolsep}{6pt}
\begin{tabular}{llrrr}
\toprule
Dataset & Policy & Mean & Median & P95 \\
\midrule
SIB-200 & Direct & .005 & .005 & .008 \\
& Fixed tier & .461 & .005 & 1.710 \\
& Paired $p$-value & .347 & .005 & 1.691 \\
& E-router & .347 & .005 & 1.691 \\
& Translate & 1.176 & 1.190 & 1.988 \\
\midrule
MASSIVE & Direct & .002 & .002 & .002 \\
& Fixed tier & .153 & .002 & .592 \\
& Paired $p$-value & .372 & .382 & .592 \\
& E-router & .372 & .382 & .592 \\
& Translate & .372 & .382 & .592 \\
\bottomrule
\end{tabular}
\end{table}

\begin{table}[ht]
\caption{Frozen e-router accuracy gain over direct-only on the held-out
confirmatory test sets. Percentile 95\% confidence intervals use 2,000 paired
bootstrap replicates with examples resampled within language.}
\label{tab:policy-accuracy-ci}
\centering
\small
\begin{tabular}{lrrr}
\toprule
Dataset & $N$ & Gain (points) & 95\% CI \\
\midrule
SIB-200 & 3,060 & 8.14 & [6.96, 9.25] \\
MASSIVE & 44,610 & 16.70 & [16.28, 17.14] \\
\bottomrule
\end{tabular}
\end{table}

\begin{figure}[p]
\centering
\includegraphics[width=.78\textwidth]{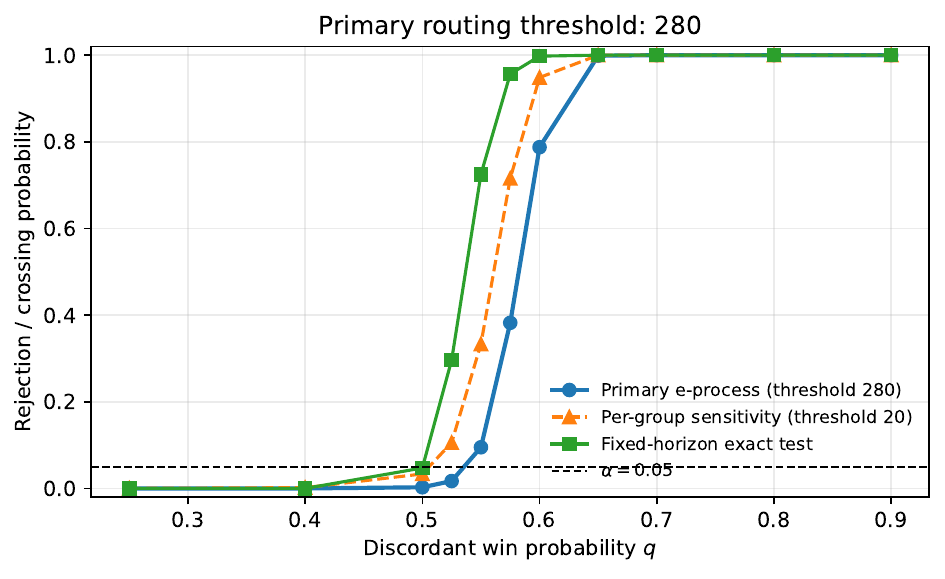}
\caption{Probability of an e-process threshold crossing by 500 discordances
under Bernoulli alternatives. The primary routing curve uses the
familywise-controlled threshold 280; threshold 20 is shown only as a per-group
sensitivity analysis. The fixed-horizon exact test is an endpoint reference.}
\label{fig:synthetic-power}
\end{figure}

\clearpage
\begin{figure}[!t]
\centering
\includegraphics[width=\textwidth]{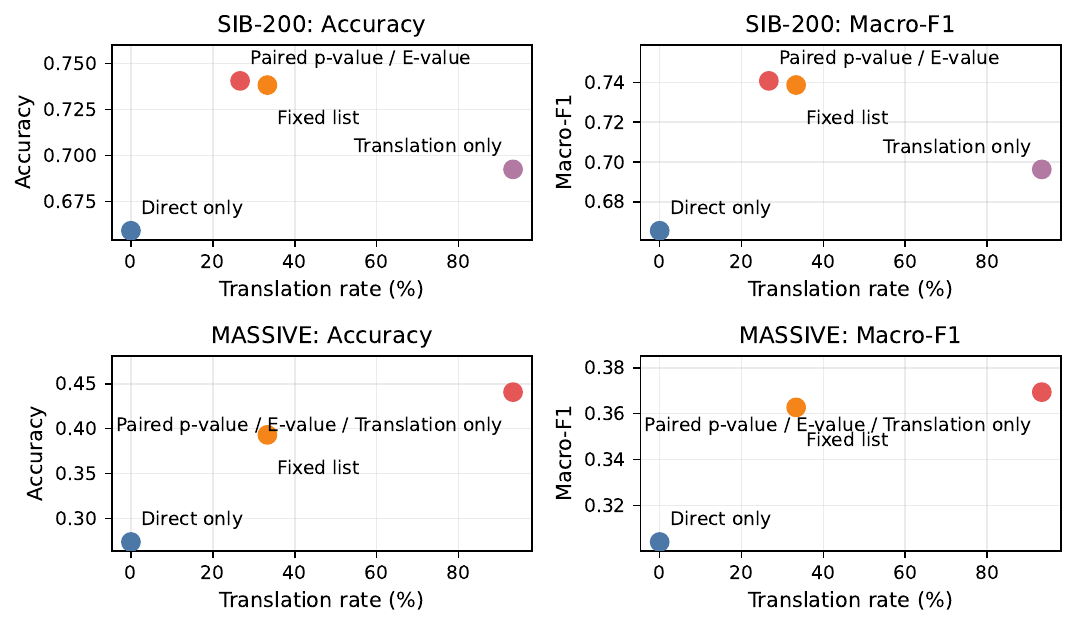}
\caption{Confirmatory quality--translation trade-off. Coincident points have
identical frozen routes.}
\label{fig:tradeoff}
\end{figure}

\clearpage
\section*{NeurIPS Paper Checklist}

\begin{enumerate}

\item {\bf Claims}
    \item[] Question: Do the main claims made in the abstract and introduction accurately reflect the paper's contributions and scope?
    \item[] Answer: \answerYes{}
    \item[] Justification: The abstract and introduction state the paper's
    methodological contribution, statistical guarantees, empirical scope, and
    task-specific conclusions consistently with Sections~2--4.

\item {\bf Limitations}
    \item[] Question: Does the paper discuss the limitations of the work performed by the authors?
    \item[] Answer: \answerYes{}
    \item[] Justification: Section~4 discusses the current accuracy-based
    target, lack of explicit utility/cost modeling, and dependence on the
    complete translation-and-classification pipeline.

\item {\bf Theory assumptions and proofs}
    \item[] Question: For each theoretical result, does the paper provide the full set of assumptions and a complete (and correct) proof?
    \item[] Answer: \answerYes{}
    \item[] Justification: Sections~2.2--2.3 state the conditional-null
    assumptions, define the filtration, justify the e-process property, and
    derive the anytime-valid crossing guarantee via the e-process maximal
    inequality.

\item {\bf Experimental result reproducibility}
    \item[] Question: Does the paper fully disclose all the information needed to reproduce the main experimental results of the paper to the extent that it affects the main claims and/or conclusions of the paper (regardless of whether the code and data are provided or not)?
    \item[] Answer: \answerYes{}
    \item[] Justification: The main paper specifies datasets, splits, model
    families, routing rules, statistical thresholds, simulation settings, and
    evaluation metrics; exact configurations and backend details are provided
    in the supplementary material and code.

\item {\bf Open access to data and code}
    \item[] Question: Does the paper provide open access to the data and code, with sufficient instructions to faithfully reproduce the main experimental results, as described in supplemental material?
    \item[] Answer: \answerYes{}
    \item[] Justification: A public code package accompanies the
    submission and includes environment specifications, configuration files,
    scripts, and instructions for reproducing the reported experiments from the
    public datasets.

\item {\bf Experimental setting/details}
    \item[] Question: Does the paper specify all the training and test details (e.g., data splits, hyperparameters, how they were chosen, type of optimizer) necessary to understand the results?
    \item[] Answer: \answerYes{}
    \item[] Justification: Section~3 specifies the datasets, discovery/test
    sizes, routing baselines, model families, statistical thresholds, and
    evaluation protocol; exact model revisions and translation backends are
    documented in the supplement.

\item {\bf Experiment statistical significance}
    \item[] Question: Does the paper report error bars suitably and correctly defined or other appropriate information about the statistical significance of the experiments?
    \item[] Answer: \answerYes{}
    \item[] Justification: Policy-level accuracy gains are reported with
    percentile 95\% confidence intervals from 2,000 paired bootstrap replicates
    with test examples resampled within language; the statistical route-discovery
    procedure is also accompanied by null-crossing and power simulations.

\item {\bf Experiments compute resources}
    \item[] Question: For each experiment, does the paper provide sufficient information on the computer resources (type of compute workers, memory, time of execution) needed to reproduce the experiments?
    \item[] Answer: \answerYes{}
    \item[] Justification: The supplement reports the local hardware,
    accelerator, memory, batching and worker settings, execution strategy, and
    recorded wall-clock time for each final discovery and evaluation run; it
    also distinguishes these runs from exploratory analyses and manual pauses.

\item {\bf Code of ethics}
    \item[] Question: Does the research conducted in the paper conform, in every respect, with the NeurIPS Code of Ethics \url{https://neurips.cc/public/EthicsGuidelines}?
    \item[] Answer: \answerYes{}
    \item[] Justification: The work uses established public benchmark datasets
    and publicly available pretrained models and does not involve human-subject
    data collection, deceptive experimentation, or restricted personal data.

\item {\bf Broader impacts}
    \item[] Question: Does the paper discuss both potential positive societal impacts and negative societal impacts of the work performed?
    \item[] Answer: \answerYes{}
    \item[] Justification: Section~4 discusses the computational-efficiency
    benefits of selective translation as well as the risk of reinforcing
    performance disparities across linguistic groups.

\item {\bf Safeguards}
    \item[] Question: Does the paper describe safeguards that have been put in place for responsible release of data or models that have a high risk for misuse (e.g., pre-trained language models, image generators, or scraped datasets)?
    \item[] Answer: \answerNA{}
    \item[] Justification: The work does not release a new high-risk model,
    scraped sensitive dataset, or other asset requiring special misuse
    safeguards.

\item {\bf Licenses for existing assets}
    \item[] Question: Are the creators or original owners of assets (e.g., code, data, models), used in the paper, properly credited and are the license and terms of use explicitly mentioned and properly respected?
    \item[] Answer: \answerYes{}
    \item[] Justification: All datasets and pretrained models are credited to
    their original sources, and their versions, licenses, and terms of use are
    documented in the supplementary material and code package.

\item {\bf New assets}
    \item[] Question: Are new assets introduced in the paper well documented and is the documentation provided alongside the assets?
    \item[] Answer: \answerYes{}
    \item[] Justification: The released experiment code is documented with
    setup instructions, configuration files, reproducibility commands, and
    descriptions of generated artifacts.

\item {\bf Crowdsourcing and research with human subjects}
    \item[] Question: For crowdsourcing experiments and research with human subjects, does the paper include the full text of instructions given to participants and screenshots, if applicable, as well as details about compensation (if any)?
    \item[] Answer: \answerNA{}
    \item[] Justification: The work uses existing benchmark datasets and does
    not conduct crowdsourcing or experiments involving human participants.

\item {\bf Institutional review board (IRB) approvals or equivalent for research with human subjects}
    \item[] Question: Does the paper describe potential risks incurred by study participants, whether such risks were disclosed to the subjects, and whether Institutional Review Board (IRB) approvals (or an equivalent approval/review based on the requirements of your country or institution) were obtained?
    \item[] Answer: \answerNA{}
    \item[] Justification: The study does not involve human-subject
    recruitment, intervention, or collection of new human-subject data.

\item {\bf Declaration of LLM usage}
    \item[] Question: Does the paper describe the usage of LLMs if it is an important, original, or non-standard component of the core methods in this research? Note that if the LLM is used only for writing, editing, or formatting purposes and does \emph{not} impact the core methodology, scientific rigor, or originality of the research, declaration is not required.
    \item[] Answer: \answerNA{}
    \item[] Justification: LLMs are not an important or non-standard component
    of the core methodology; the study evaluates pretrained multilingual and
    English encoders and machine-translation models.

\end{enumerate}

\end{document}